\documentclass[11pt,a4paper]{article}
\usepackage[hyperref]{emnlp2020}
\usepackage{times}
\usepackage{latexsym}
\usepackage{amsmath}
\usepackage{graphicx}
\usepackage{subcaption}
\usepackage{nameref}
\DeclareMathOperator*{\argmax}{arg\,max}

\usepackage{array}
\newcolumntype{R}[1]{>{\raggedleft\let\newline\\\arraybackslash\hspace{0pt}}m{#1}}

\usepackage{microtype}

\usepackage{soul}
\setuldepth{o}

\aclfinalcopy 

\title{Analogies minus analogy test: measuring regularities in word embeddings
}

\author{Louis Fournier $^1$ \\
   \\
   \\
   \\
   \\
  \ \ \ \ \ \ \ \ \ \ \ \ \ \ \ \ \ \  \texttt{louis.fournier@polytechnique.edu} \\
  \And
  Emmanuel Dupoux $^1$ $^2$ \\
  $^1$ Cognitive Machine Learning (ENS - CNRS - EHESS - INRIA - PSL Research University), France \\
  $^2$ Facebook A.I. Research, Paris, France \\
  $^3$ Laboratoire de Linguistique Formelle (CNRS - Paris Diderot - Sorbonne Paris Cite), France  \\
  $^4$ University of Toronto, Toronto, Canada \\
\\
  \texttt{emmanuel.dupoux@gmail.com} \ \ \ \ \ \ \ \ \  
  \And
  Ewan Dunbar $^1$ $^3$ $^4$\\

  \\
     \\
   \\
   \\
  \texttt{ewan.dunbar@utoronto.ca} \\
  }

\date{}

\begin{document}
\maketitle
\begin{abstract}
Vector space models of words have long been claimed to capture linguistic regularities as simple vector translations, but problems have been raised with this claim. We decompose and empirically analyze the classic arithmetic word analogy test, to motivate two new metrics that address the issues with the standard test, and which distinguish between class-wise \textbf{offset concentration} (similar directions between pairs of words drawn from different broad classes, such as \emph{France--London}, \emph{China--Ottawa,} \dots) and \textbf{pairing consistency} (the existence of a regular transformation between \emph{correctly}-matched pairs such as \emph{France:Paris::China:Beijing}). We show that, while the standard analogy test is flawed, several popular word embeddings do nevertheless encode linguistic regularities. 

\end{abstract}

\section{Introduction}

Vector semantic models
saw a surge in interest after embeddings trained under the word2vec architecture \cite{MikolovChenEtAl_2013_Efficient_estimation_of_word_representations_in_vector_space} 
were shown to encode linguistic regularities  \cite{MikolovYihEtAl_2013_Linguistic_Regularities_in_Continuous_Space_Word_Representations}.
The demonstration relied on the arithmetic analogy test: relations such as $\textrm{king}\!+\!\textrm{woman}\!-\!\textrm{man}\!\approx\!\textrm{queen}$ were shown to hold for a variety of semantic and grammatical relations.
Evaluation of word embeddings on analogy tests and training on related loss functions remains current. There is also continued interest in 
theoretically grounding the 
success of distributional embeddings on these tests   \citep{allen,ethayarajh-etal-2019-towards}. 


 There is, however,  a substantial literature pointing to problems with word analogies (see Section \ref{sec:related-work}), leading to the conclusion that word analogies are fragile in practice, and sometimes going so far as to imply that the positive results were erroneous. These critiques have been ambiguous as to whether the problem is that the embeddings in question do not really encode the relevant linguistic structure, or whether the issue is merely that the arithmetic analogy \emph{test} as it is standardly defined, is flawed. 
 
 The current paper confirms that there are serious problems with the standard analogy test, as it confounds three different properties of word vectors while purporting to measure only one: class-wise \textbf{offset concentration} (similar directions between pairs of words drawn from different broad classes, such as \emph{France--London}, \emph{China--Ottawa}), \textbf{within-pair similarity} between test words (for example, the similarity between \emph{Paris} and \emph{France}), as well as the \textbf{pairing consistency} the test sets out to measure (the presence of a regular direction encoding relations such as \emph{capital-of}:  \emph{France--Paris},  \emph{China--Beijing}). We give an algebraic decomposition of the standard analogy test that explains previous negative results in terms of within-pair similarity. Using new measures, we show that, in practice, offset concentration, rather than true pairing consistency, may account for part of word embeddings' success. Nevertheless, we show that several standard word embeddings do show pairing consistency for the relations tested by the BATS analogy benchmark \cite{GladkovaDrozd2016}.\footnote{Code is available at \url{github.com/bootphon/measuring-regularities-in-word-embeddings}.}

\section{Related work}\label{sec:related-work}

The validity of the arithmetic analogy test has been questioned in several papers, starting with \citet{levy-goldberg-2014-linguistic}. We detail in Section \ref{sec:analysis} several major issues with the test as raised by \citet{levy-goldberg-2014-linguistic}, \citet{Linzen_2016_Issues_in_evaluating_semantic_spaces_using_word_analogies}, and \citet{RogersDrozdEtAl_2017_Too_Many_Problems_of_Analogical_Reasoning_with_Word_Vectors}. \citet{finley-etal-2017-analogies}, \citet{newman-griffis-etal-2017-insights}, \citet{chen-2017} and \citet{schluter-2018-word} also raised  concerns about the  test and its assumptions. 
More recently, \citet{fairisbetter} argued against this test as an inadequate tool for studying bias in word embeddings. \citet{Rogers_2019_analogies} observes that many of the issues have been ignored.
Some works have proposed other measures of linguistic relations in word embeddings.
\citet{levy-goldberg-2014-linguistic}, \citet{vylomova-etal-2016-take}, and \citet{RogersDrozdEtAl_2017_Too_Many_Problems_of_Analogical_Reasoning_with_Word_Vectors} all examined the similarity of vector offsets in a more direct way than the standard analogy test (see Section \ref{sec:analysis} below).
\citet{DrozdGladkovaEtAl_2016_Word_embeddings_analogies_and_machine_learning_beyond_king_man_woman_queen} proposed a method based on 
predicting the class of the unknown word, and \citet{bouraoui-etal-2018-relation} 
relaxed the assumptions by  allowing probabilistic models to predict the relations. 







We claim that these works still do not provide a satisfactory measure of how well word vector offsets encode linguistic relations. Without such a measure, it is impossible to assess whether the original conclusions are correct. We develop this argument, and then develop a new measure, below.

\section{The arithmetic analogy test}\label{sec:analysis}

\citet{MikolovYihEtAl_2013_Linguistic_Regularities_in_Continuous_Space_Word_Representations}
proposed to measure the presence of linguistic relations in the structure of word embeddings using vector arithmetic. For two pairs of words representing the same linguistic relation, $(a,a^*)$ and $(b,b^*)$, the test assesses whether: 
\setlength{\abovedisplayskip}{7pt}
\setlength{\belowdisplayskip}{8pt}
\begin{equation}
b + (a^* - a) \approx b^*
\end{equation}

In practice, the test assesses whether $b^*$ is the nearest neighbour to $b + a^* - a$, using the cosine similarity (cosine of the angle between  $x$ and $y$): 
\setlength{\abovedisplayskip}{7pt}
\setlength{\belowdisplayskip}{8pt}
\begin{equation}
    \text{sim}(x, y)  = \frac{
    x\cdot y}{\left\lVert x\right\rVert\!\left\lVert y\right\rVert}
\end{equation}

The assumption is that, if the offset $o_a\!=\!a^*\!-\!a$  is parallel to the offset $o_b\!=\!b^*\!-\!b$, then this common offset direction encodes the same linguistic relation. For example, if \emph{Paris} minus \emph{France} goes in the same direction as \emph{Beijing} minus \emph{China}, and as other capital--country pairs,  the common offset direction can be seen as encoding the relation \emph{capital-of.} 

Thus, the idea behind the arithmetic analogy test 
is that  offsets corresponding to the same linguistic relation, if they are the same or roughly the same, should be interchangeable. Therefore, \emph{France}~$+$~(\emph{Beijing}~$-$~\emph{China}) should be the same as \emph{France}~$+$~(\emph{Paris}~$-$~\emph{France}) (modulo the vector magnitude, if we are hypothesizing that it is the direction, and not the precise normed vector,  that encodes the relation \emph{capital-of:} hence the use of the cosine similarity). 

The criticisms of \citet{MikolovYihEtAl_2013_Linguistic_Regularities_in_Continuous_Space_Word_Representations} and the research it inspired have often been characterized as problems with ``word analogies.'' 
This is ambiguous: it does not distinguish between problems with the \emph{method} of using the arithmetic test just described for testing the presence of linguistic regularities, and the veracity of the \emph{conclusion} that there are linguistic regularities coded in word embeddings as vector offsets. In order to resolve this ambiguity, one would need to know for sure, using some better measure, whether linguistic regularities are coded. We propose such a measure beginning in Section \ref{sec:ocspcs}. We first analyze the key problems that have been previously raised. 





\subsection{Within-pair similarity}

The analogy will be scored correct if $b^*$ is the $\argmax_x{\text{sim}(b+o_a, x)}$.  \citet{levy-goldberg-2014-linguistic} call this objective \textsc{3CosAdd}. Let us call $\text{sim}(b+o_a, b^*)$ the ``analogy score.''  Putting aside  norms, the analogy score can be decomposed as: 
\setlength{\abovedisplayskip}{7pt}
\setlength{\belowdisplayskip}{8pt}
\begin{equation}\label{eq:decomp-analogy-score}
\begin{split}
\text{sim}(b+o_a, b^*) & \propto  b\cdot b^* + o_a\cdot b^*\\
& \propto b\cdot b^* + o_a\cdot o_b + o_a\cdot b
\end{split}
\end{equation}





 The first term, the dot product of  $b$ and $b^*$,  is proportional to the \textbf{within-pair similarity}, $\text{sim}(b, b^*)$. The second term is proportional to the similarity between the offsets, which would appear at first glance, to be the term of principal interest in assessing whether a linguistic relation is consistently coded by the offsets. 
 The third term is proportional to the similarity between the start word $b$ and the offset $o_a$, and does not depend on $b^*$. 



We develop an analysis of the offset similarity in sections \ref{sec:ocspcs}--\ref{sec:method-ocspcs}, deriving from it the offset concentration and pairing consistency properties we propose to measure. Previous work suggested that offset similarity is not very high in practice, compared to within-pair similarity. \citet{levy-goldberg-2014-linguistic} showed that evaluating on the basis of similarity between offsets, using \textsc{PairDirection} to determine the nearest neighbours, leads to failure on analogy tests that would otherwise succeed using the \textsc{3CosAdd} objective.  \citet{RogersDrozdEtAl_2017_Too_Many_Problems_of_Analogical_Reasoning_with_Word_Vectors} showed that within-pair similarity is correlated with performance on the standard arithmetic analogy test. Clearly, measuring whether \emph{Paris} is similar to \emph{France} is not the same as measuring whether the relation \emph{capital-of} is present in the word embedding. As \citeauthor{levy-goldberg-2014-linguistic} point out, within-pair similarity is not irrelevant to the question---if the idea is that there is a consistent dimension \emph{capital-of} which should constitute the principal difference between \emph{Paris} and \emph{France}, then the two words should otherwise be similar.  But a test dominated by within-pair similarity can lead to spurious results, since pairs of words may be similar without capturing the relation in question. 

Using the Bigger Analogy Test Set (BATS)  \cite{GladkovaDrozd2016} and the pre-trained Google News skip-gram embeddings of \citet{MikolovChenEtAl_2013_Efficient_estimation_of_word_representations_in_vector_space}, we examine the role of the three terms of \eqref{eq:decomp-analogy-score}. BATS consists of forty relations, each composed of fifty word pairs, grouped into four broad types: inflectional morphology, derivational morphology, encyclopedic semantics (like capitals and animal sounds), and lexicographic semantics (like synonyms and antonyms). When multiple pairs exist for the same input word, we only keep the first. For each of the forty relations, we take the mean for each term over all $a\!:\!a^*\!::\!b\!:\!b^*$ tuples. 

\begin{figure*}[t]
  \centering
\includegraphics[width=.9\linewidth]{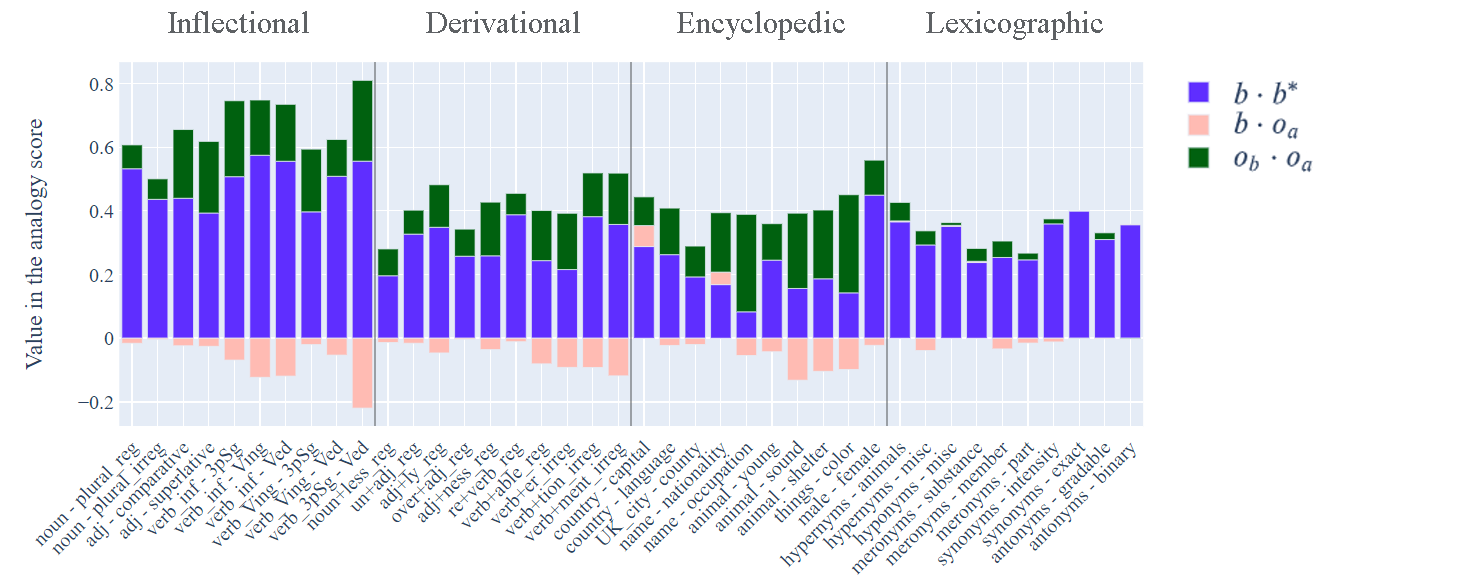}
  \caption{Decomposition of the  analogy score in three terms, with only two depending on the word to predict.  The  largest component is $b\cdot b^*$,  proportional to  within-pair similarity.  $b\cdot o_a$ is proportional to the similarity of the offset to the start word $b$. $o_b\cdot o_a$ is proportional to  the similarity between offsets.  All terms are divided by the same overall normalization term (not indicated in the legend).} 
  \label{fig:bar decompo}
\end{figure*}

Complementing previous analyses, we display  means for these three terms in Figure \ref{fig:bar decompo}, confirming that within-pair similarity is indeed positive and large compared to the other terms. Notably,  offset similarities vary between relations, but are almost always smaller. However, they are always positive. 

\subsection{Honest analogies: predicting input words}

\citet{Linzen_2016_Issues_in_evaluating_semantic_spaces_using_word_analogies} observes that the arithmetic analogy test, as practised, excludes $a$, $a^*$, and $b$ from the $\argmax$---otherwise, one of these input words tends (incorrectly) to be the response---most of the time, $b$. \citet{RogersDrozdEtAl_2017_Too_Many_Problems_of_Analogical_Reasoning_with_Word_Vectors} and \citet{schluter-2018-word} imply that part of the issue may be that $\|o_a\|$ is small.
This cannot be why: the offsets have similar magnitudes,  so if $\left\lVert o_b\right\rVert$ is long enough to bring $b$ to $b^*$, then $\left\lVert o_a\right\rVert$ should be too. 
Importantly, if we compute the decomposition in \eqref{eq:decomp-analogy-score}, but suppose that $b^*$ is  equal to $b+o_a$ ($o_a\!=o_b$: the analogy score is thus always 1), we observe empirically that the within-pair similarity-  and offset similarity-driven terms have similar values (Figure 8 
in the appendix). This means that $\left\lVert o_a\right\rVert$
is similar to $\|b\|$. The weakness of the offset-similarity term in the score is not driven by the offsets being small.

Define $\Delta_\text{sim}$ as the analogy score minus the similarity of $b + o_a$ to the start word $b$, which must be negative in order for the ``honest'' analogy test to return $b$. We observe that $\Delta_\text{sim}$ is equal to:
\setlength{\abovedisplayskip}{7pt}
\setlength{\belowdisplayskip}{8pt}
\begin{equation}
 \Delta_\text{sim} = \frac{b\!+\!o_a}{\|b\!+\!o_a\|}\!\cdot\!(\frac{b^*}{\|b^*\|}\!-\!\frac{b}{\|b\|}) 
\end{equation}

By replacing $b^*$ by $b+o_a$, we can expand the score, which is then proportional to:
\setlength{\abovedisplayskip}{7pt}
\setlength{\belowdisplayskip}{8pt}
\begin{equation}\label{eq:decomp-delta}
\frac{\|b\|\!-\!\|b^*\|}{\|b\|}\!\cdot\!(b\!+\!o_a)\!\cdot\!b  + o_a\!\cdot\!o_b + b\!\cdot\!o_b
\end{equation}

This decomposes $\Delta_\text{sim}$  into three terms. The first term is proportional to the difference between the norms of $b$ and $b^*$, which we have empirically observed to be small, and which will be null for normalized embeddings. The second term is proportional to the offset similarity, and the third term is proportional to the similarity between $b$ and $o_b$. For normalized word vectors, the third term is equal to $\text{sim}(b,b^*)-1$, which is the  negative cosine distance between $b$ and $b^*$. Since $\|b\|$ and $\|b^*\|$ tend to be similar, we find that this is a good approximation even for non-normalized vectors. $\Delta_\text{sim}$ is thus negative, roughly, whenever the offset similarity is less than the cosine distance between $b$ and $b^*$. Intuitively, the similarity of the offsets must compensate for the difference between $b^*$ and $b$. 

\begin{figure*}[ht]
\centering
\includegraphics[width=.9\linewidth]{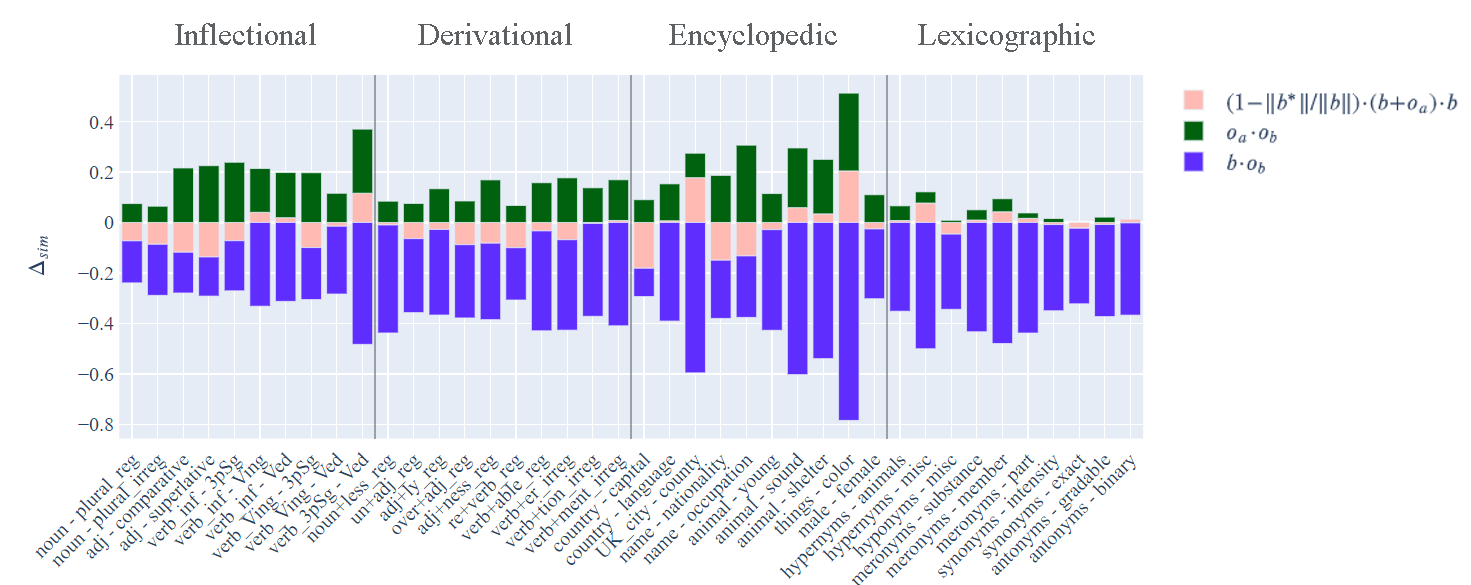}
\caption{Decomposition of $\Delta_\text{sim}$, the difference between the similarity of the analogy to $b^*$ and to $b$. The term proportional to the dissimilarity of $b$ to $b^*$ is greater than the one proportional to the offset similarity, which means the common direction of the offset is not enough to bring the analogy closer to $b^*$ than to $b$.}
\label{fig:bar diff sim}
\end{figure*}

In Figure \ref{fig:bar diff sim}, we plot means for the three terms from \eqref{eq:decomp-delta}. We observe that the first term  is indeed small and can be ignored. Importantly, we observe that the offset similarity term  is smaller in magnitude than the negative $b$--$b^*$  distance term). 

\citeauthor{Linzen_2016_Issues_in_evaluating_semantic_spaces_using_word_analogies}'s observation that an input word is typically predicted under the \textsc{3CosAdd} objective---and not an unrelated vocabulary word---implies that not only is $\Delta_\text{sim}$ negative, but that the equivalent value is negative for all words in the vocabulary. This in turn would imply (for normalized embeddings) that the similarity between $o_a$ and the offset of $b$ with respect to any other word is smaller than the distance between $b$ and that word. Far from staying in the neighbourhood of $b$, $o_a$ moves $b$ in a direction far from every word in the vocabulary. 

\begin{figure}\centering
\includegraphics[width=\linewidth]{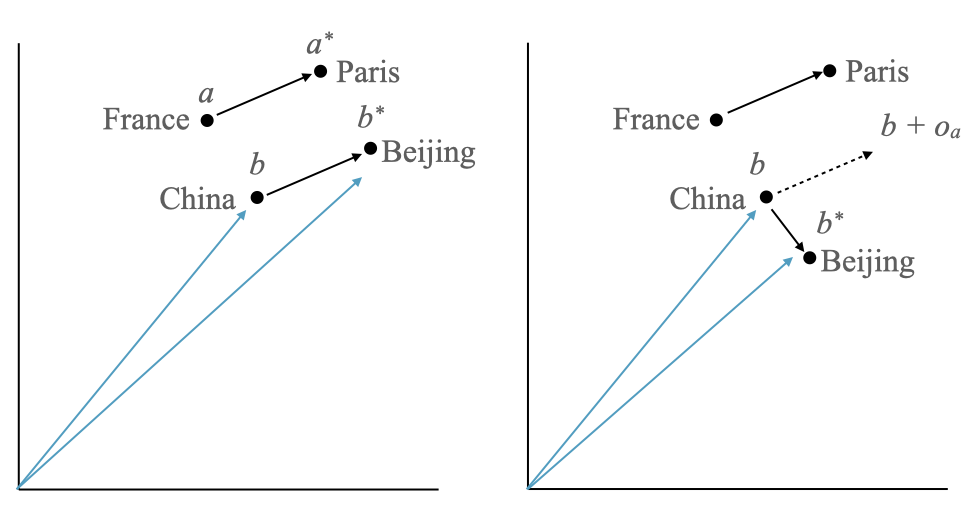}
\caption{Schematic illustration of an analogy underlining the role of within-pair similarity in the arithmetic analogy test.
At left, \emph{China}~$+$~$($\emph{Paris}~$-$~\emph{France}$)$ predicts \emph{Beijing} in part for the ``right'' reason (offsets are parallel). At right, the offsets are not parallel, but 
\emph{Beijing} is likely to be predicted, since its cosine similarity to \emph{China} is high. The problem is worse than the 2D schematic would suggests, since it can occur even when the word vectors are normalized, and when the offsets are quite long. \emph{China} will be predicted if it is not excluded, since it is also in the neighbourhood.  
}
\label{fig:analogy drawing}
\end{figure}

Our analysis of these known problems with word analogies  details the central problem: the pairs of offsets tested are not similar enough to support the kind of arithmetic demanded by the \textsc{3CosAdd} objective. An illustration of the dominance of within-pair similarity is given in Figure \ref{fig:analogy drawing}. Note that we present this figure \emph{only} to illustrate the dominance of within-pair similarity. Two-dimensional projections are misleading: the vectors in Figure \ref{fig:analogy drawing} do not have the property, discussed above, that the offsets are of length similar to the word vectors themselves---this  is difficult to reconcile with high within-pair similarity in two dimensions. 


None of this, however, implies that word embeddings do not show linguistic regularities. It simply implies that the standard arithmetic analogy test does not  measure these regularities correctly, and should not be used. While it may be true that the quantity of  interest, the similarity between offsets, is generally insufficient to dominate the standard analogy test, that does not mean it is too small to be considered an encoding of a linguistic regularity. How similar is similar enough? In the following section, we propose an answer to this question.


\section{Measuring Regularities}\label{sec:ocspcs}

The similarity between offsets, $\text{sim}(o_a,o_b)$, is the quantity of primary interest to assess whether linguistic relations are represented as consistent directions. We seek a measure of whether they are.
As noted, \citet{levy-goldberg-2014-linguistic} showed that   $\argmax_x{\text{sim}(a^*-a, x-b)}$ does not tend to find the right answer ($b^*$).  However, the fact that $b^*$ does not maximize offset similarity does not answer the question of what a meaningful level of offset similarity is. \citet{RogersDrozdEtAl_2017_Too_Many_Problems_of_Analogical_Reasoning_with_Word_Vectors} analyzed examples of correct and false analogies and showed them to have comparable offset similarities. This makes sense: a baseline level of offset similarity can be found by looking at unrelated pairs. However, that paper examined only one isolated example.

We propose an \textbf{offset concentration score} (OCS) measuring the  offset similarity among  pairs belonging to the same relation and a \textbf{pairing consistency score} (PCS) measuring the distinctness of the similarities from a baseline. We propose  PCS as the correct measure of linguistic regularity, with OCS a  source of supplementary information.

\emph{Perfectly} parallel offsets would imply that \emph{France} and \emph{Paris} differ on the same dimension as \emph{China} and \emph{Beijing}---\emph{and on no other dimension}---and similarly for the other offsets of the same kind (\emph{Canada}$-$\emph{Ottawa}, \dots).
 Perfect parallelism is not necessary for  $\text{sim}(o_a,o_b)$ to contribute to success on the analogy test.  By offset concentration, we mean the degree to which, for an entire set of word pairs, the offsets are parallel to one another (thus, concentrated in a single direction). 
 
\begin{figure}\centering
\includegraphics[width=\linewidth]{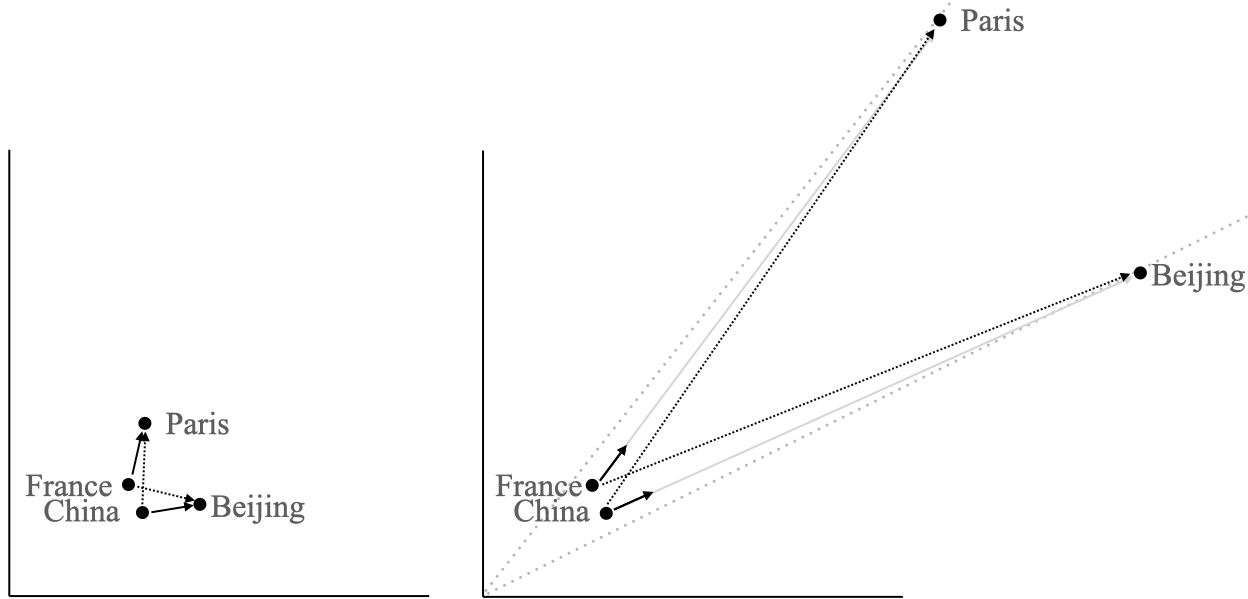}
\caption{Schematic illustration of an analogy underlining the pitfalls of offset concentration.
At left, \emph{Paris}~$-$~\emph{France} appears nearly orthogonal to \emph{Beijing}~$-$~\emph{China}. At right, however, the offsets have much greater cosine similarity, simply because \emph{Paris} and \emph{Beijing} are far away.  The similarity between \emph{Paris}~$-$~\emph{China} and \emph{Beijing}~$-$~\emph{France} (dotted lines) is almost as high. Unlike what might be suggested by the 2D schema, this can arise easily even for normalized word vectors.   
}
\label{fig:ocs-schema}
\end{figure}

 Once we take into account that offsets need not be perfectly parallel, we must bear in mind that positive cosine similarity between offsets does not imply that the embedding space captures linguistic relations. For example, recalling \citet{schluter-2018-word}, training embeddings to capture distributional regularities may group words into  linguistic classes---with, for example, country names such as \emph{France}, \emph{China}\dots, occupying a region of the space  distinct from that of \emph{Paris}, \emph{Beijijng}\dots, due to distributional commonalities. This by itself could yield offset concentration: offsets would all come from a similar origin, and go to a similar destination. 
 However,  $\text{sim}(\mbox{\emph{France}}$-$ \mbox{\emph{Paris}}, \mbox{\emph{China}}$-$\mbox{\emph{Beijing}})$ would not necessarily be larger than $\text{sim}(\mbox{\emph{France}}$-$ \mbox{\emph{Beijing}}, \mbox{\emph{China}}$-$\mbox{\emph{Paris}})$. We illustrate this issue in Figure \ref{fig:ocs-schema}. 

 In such an example, we might be able to assert that the class of capital city names is distinct from the class of country names. But asserting that the relation \emph{capital-\textbf{\ul{of}}} is captured is a stronger property. It would require that moving from  the \emph{France} vector in the direction of the common \emph{capital-of} vector lead to \emph{Paris} and not to \emph{Beijing}, nor to any other capital city---or, at least, that it passes closer to \emph{Paris.}  We call this property pairing consistency.


\subsection{Offset concentration}

\begin{figure*}[t]
\centering
\includegraphics[width=.8\linewidth]{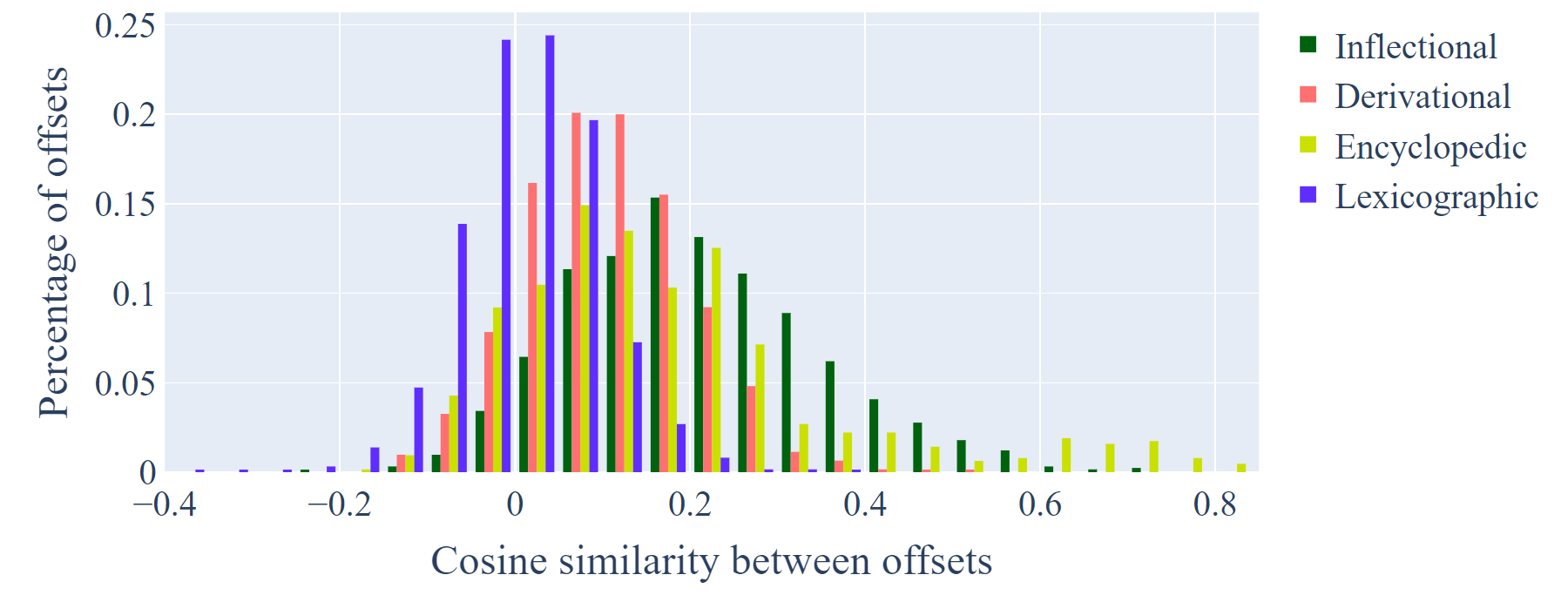}
\caption{Distribution of the cosine similarities between vector offsets for word2vec embeddings,  grouped by broad relation type. The Offset Concentration Score is the mean offset similarity for all pairs within a single relation (e.g., \emph{country}--\emph{capital}).
}
\label{fig:distrs}
\end{figure*}

Figure \ref{fig:distrs} shows a histogram of the pairwise similarities between all pairs of BATS analogy offsets for the pre-trained word2vec embeddings used above, grouped by broad BATS relation type. As above,  these similarities  are small but generally positive. We propose an  \textbf{offset concentration score}:
\setlength{\abovedisplayskip}{7pt}
\setlength{\belowdisplayskip}{8pt}
\begin{equation} 
\textrm{OCS} = \frac{1}{N(N-1)}\sum_i \sum_{j\neq i} o_i \cdot o_j
\end{equation}
which is the mean similarity between normalized offsets within a single relation, where the normalized offset vector $o_i$ of a given word pair  is $o_i\!=\!\frac{a_i^*\!-\!a_i}{\|a_i^*\!-\!a_i\|}$. 
We discuss in the appendix 
the link between the OCS and the mean direction of the offsets. 

\subsection{Pairing consistency\label{scores}}

To measure the pairing consistency, we compare the direction of the offsets for a given set of word pairs against comparable 
shuffled offsets from the same BATS  relation. Shuffled offsets are generated by holding the start words constant but permuting the end words (avoiding true pairs). We expect true offsets to be more parallel than shuffled offsets. 

We sample sets of shuffled offsets for each relation. For each set, as for the  true offsets, we compute all pairwise similarities between offsets. Rather than averaging similarities as in the OCS, we calculate the separation between  distributions---the true offset similarities versus each set of  shuffled offset similarities. We empirically construct the receiver operating characteristic curve for linearly separating true versus shuffled similarities (i.e., setting a  criterion on the similarity dimension above which  offsets are true offsets and below which  offsets are shuffled offsets). We  calculate the area under the curve (AUC). 
The AUC will be 1 for perfectly separable distributions, 0.5 for indistinguishable distributions, and below 0.5 when true offsets are systematically less parallel than the baseline. We average  AUC over all shuffled sets.  

For any set of offsets $O$ 
, we refer to the set of its 
offset similarities as $\text{sim}(O) = \{ o_i\cdot o_j, \forall (o_i,o_j)\in O^2, i<j \}$. 
For $S_s$ a list of $N_s$  sets of shuffled offsets  $O^\prime$ (here $N_s=50$),
we define the following \textbf{pairing consistency score}, where $O$ are the true offsets: 
\setlength{\abovedisplayskip}{7pt}
\setlength{\belowdisplayskip}{8pt}
\begin{equation} 
\textrm{PCS} = \frac{1}{N_s}\!\sum_{O^\prime\in S_s}\!\textrm{AUC}\big(\text{sim}(O) , \text{sim}(O^\prime)\big)
\end{equation}

\section{Exploration of word2vec}\label{sec:method-ocspcs}

We perform a series of experiments to assess the presence of linguistic regularities in the pre-trained word2vec embeddings used above, and to show the behavior of our new measures.

\subsection{Random baselines}

In addition to the real BATS test, we apply our measures to artificial analogy sets, which we construct to show  no pairing consistency. We seek to justify the use of pairing consistency, rather than directly using offset concentration, by demonstrating that offset concentrations can remain non-zero even for arbitrarily associated pairs.
For each BATS relation, we construct ten of each of the following types of permuted relations. We calculate the mean OCS and the mean PCS over the ten instances. We expect the mean PCS to be 0.5 in each case. 

First, we construct sets of word pairs derived from the BATS by \textbf{permutation within-category}: for each category, the end words are randomly re-assigned to start words. In other words, we construct categories for which shuffling should have no effect (they are already shuffled). Comparing the OCS to that of the real BATS also allows us to measure the effect of shuffling on the OCS. 

If permuted word pairs can show positive offset concentration simply because the start words and/or the end words are found in a coherent region of the embedding space, then positive OCS should be found in word pairs drawn from \textbf{mismatched categories}. For each pair of start/end word categories among the BATS relations (for example, the \emph{noun}--\emph{plural\_reg} relation maps the start category \emph{singular nouns} to the end category \emph{regular plural nouns}), we randomly re-assign the end category (yielding, for example, \emph{singular nouns}--\emph{past tense verbs}). We then randomly construct fifty start/end pairs. 
We compare mismatched categories from \textbf{within} versus \textbf{across} broad BATS category types (inflectional morphology, derivational morphology, encyclopedic semantics, lexicographic semantics).

Finally, to assess the impact of the geometric coherence of words within categories on the OCS, we construct pairs without respect for category: we compare word pairs with \textbf{random start} words, \textbf{random end} words, and \textbf{random start and end} words, with ten different random word sets.

\subsection{Offset concentration \label{simr}}

\begin{table*}
\centering
\scalebox{.9}{\begin{tabular}{l||p{.95cm}p{.95cm}|p{.95cm}p{.95cm}|p{.95cm}p{.95cm}|p{.95cm}p{.95cm}}
 & \multicolumn{2}{c|}{\textbf{Inflectional}}& \multicolumn{2}{c|}{\textbf{Derivational}}   & \multicolumn{2}{c|}{\textbf{Encyclopedic}}&  \multicolumn{2}{c}{\textbf{Lexicographic}} \\ 
\textbf{Analogy set} &  OCS & PCS & OCS & PCS & OCS & PCS & OCS & PCS\\ \hline 
Real BATS & \textbf{.295} & {.851}  &\textbf{.156} & {.679}  & .198 & {.559}&  .031 & {.539} \\\hline 
Permuted within-category & .111 & .500  & .088 & .500  & .170 & .500 &   .015 & .500 \\
Mismatched category (within type) & .147 & .501 &.120 & .500 & \textbf{.260} & .501  &  .093 & .500 \\
Mismatched category (across type) & .175 & .500 &.173 & .500 & .223 & .500 &  \textbf{.134} & .500 \\\hline 
Random start &  .090 & .500 & .075 & .500 & .204 & .499 & .069 & .500  \\
Random end & .063 & .500  &.060 & .500 & .137 & .501 &  .059 & .499\\\hline
Random start and end & .000 & .500  & \multicolumn{6}{r}{ } \\
\end{tabular}}
\caption{Offset Concentration Scores and Pairing Consistency Scores for the real and random baseline analogy sets (higher is better; 0.5 is chance level for PCS). Random baseline scores are averaged across ten permutations.  All scores are then averaged across BATS relations. Mismatched categories (across broad BATS relation type) are grouped along the start  type. Random start and end is not assigned to a BATS type, as all the words are drawn randomly. Bold indicates the highest OCS score in a given column: we see that OCS is not always highest for real analogies.  Random baseline PCS scores are always within a half-IQR of 0.5.}
\label{tab:baseline sets}
\end{table*}

Table \ref{tab:baseline sets} shows the OCS and PCS scores on real and randomized BATS relations using word2vec. The OCS scores for the real BATS sets are not close to one (as expected, given the results above). 

The fully random set has an OCS of zero, confirming that offset concentration is unlikely to occur by chance, at least for these word vectors. The other random analogy sets all show non-zero OCS. Some random sets have even higher OCS scores than the real BATS sets. Thus, as predicted, offset concentration can exist even when linguistic relations are, by construction, not coded in any systematic way.
Even the random-start and random-end pairs show non-zero OCS,\footnote{We observe a systematic difference between the random start and random end results, which seems to indicate a bias in the BATS analogy sets, with the end words being more susceptible to forming clusters.} indicating that geometric coherence within one category is sufficient to give some degree of parallelism to the offsets.  
Surprisingly, the OCS is systematically lower for the permuted within-category baseline than for the mismatched-category baselines. Mismatched categories may be further apart than matched categories, reducing the angles between offsets overall.  

These results have important consequences for how we measure meaningful regularities in vector embeddings. Not only is the arithmetic analogy test not fit to detect the presence of similar directions among related pairs, even if it were,  similar directions alone are not sufficient to deduce that linguistic regularities are coded. The test can be dominated by spurious within-pair similarity \emph{and} by spurious offset concentration---which can arise even when matching against random words. 

\subsection{Pairing consistency \label{scoresr}}

The real BATS sets show PCS above chance, contrary to all  random analogy sets, which show PCS  of 0.5, as expected.\footnote{As noted, these scores are the mean over ten replications of the random sets. Even where the mean is not precisely 0.5, it is always within one half-interquartile-range of 0.5.} The PCS scores are low for the  semantic (encyclopedic, lexicographic) relations. 


In Section \ref{sec:analysis} above, we remarked that 
within-pair similarity is not completely irrelevant to analogies, as a higher within-pair similarity  amplifies the component along the common direction. 
Here, the relations which show greater within-pair similarity (Figure \ref{fig:bar decompo}) also show higher PCS.

\section{Comparing word embeddings} 


We evaluate the pairing consistency on the BATS test for popular word embeddings. 
We include Glove \cite{pennington-etal-2014-glove},  a purely distributional model similar to word2vec; two embeddings making use of external semantic knowledge: dict2vec \cite{tissier-etal-2017-dict2vec} and ConceptNet Numberbatch \cite{speer2017conceptnet}; and the static token embeddings of BERT \cite{devlin-etal-2019-bert} and GPT-2 \cite{radford2019language}. 

\begin{figure*}[t]
\centering
\includegraphics[height=1.58in]{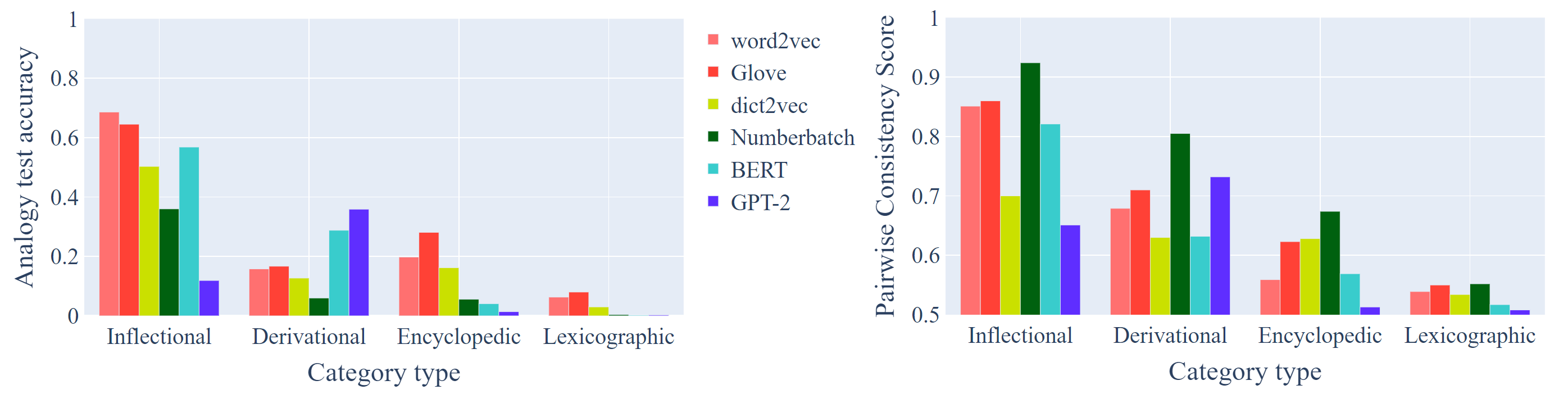}

\caption{Left: accuracy on the standard arithmetic analogy test, across embeddings, averaged across BATS relations (within the four broad types). Right: the same figure for the Pairing Consistency Score, starting at .5 (chance-level).}
\label{fig:bar-emb-pcs}
\end{figure*}

The left subfigure in Figure \ref{fig:bar-emb-pcs} presents the standard  \textsc{3CosAdd} analogy test accuracy (input words excluded). Within broad relation types, we see  similar results across embeddings, with a few exceptions (the static GPT-2 embeddings are poor throughout, but much better than the others on derivational morphology). Consistent with previous results using the arithmetic analogy test, derivational morphology and encyclopedic and lexical semantic relations obtain lower scores than inflectional morphology \citep{GladkovaDrozd2016,RogersDrozdEtAl_2017_Too_Many_Problems_of_Analogical_Reasoning_with_Word_Vectors}.

The PCS scores are presented at right in Figure \ref{fig:bar-emb-pcs}. The general picture is similar: there are systematic differences between relation types, and inflectional morphology tends to be well coded.  However, contrary to the conclusions of the standard analogy test, the PCS reveals that Numberbatch codes  linguistic relations better than the other embeddings, which is more consistent with the evaluations presented in \citet{speer2017conceptnet}. The PCS also shows that derivational morphology  occupies an intermediate position between inflectional morphology and semantics:  word2vec, Glove, and Numberbatch in fact
code derivational morphology better, in the PCS sense, than encyclopedic semantic relations. The elevated analogy test performance of some embeddings on encyclopedic relations is likely an artefact of the kind of category-level geometric coherence discussed in Section \ref{sec:method-ocspcs}. As we show in the appendix in Table 2 
, encyclopedic relations show higher OCS overall than derivational morphology relations, in spite of lower PCS.

\section{Contributions}

We have made new arguments against the use of the standard arithmetic analogy test in the evaluation of word embeddings. We show in detail 
how regularities in vector offsets, which are the object of primary interest for the analogy test, are, on the one hand, washed out by 
within-pair similarity, and, on the other, insufficient to establish the existence of linguistic regularities. We 
explain the previously observed phenomenon that 
arithmetic  analogy tests 
consistently predict the input words. 

We propose a new measure of the presence of linguistic relations in word embeddings, the pairing consistency score, which measures the degree to which offsets between related pairs are parallel above chance, and the offset concentration score, a complementary measure of the absolute degree of parallelism.
We show that a variety of word vectors really do capture linguistic regularities, in spite of  distortions introduced by the arithmetic analogy test. We show that these distortions lead to spurious conclusions when comparing performance. Arithmetic analogy tests and loss functions based on them are deprecated and should be replaced by  direct evaluation of pairing consistency.

\section*{Acknowledgments}

This work was funded in part by the European Research Council (ERC-2011-AdG-295810 BOOTPHON), the Agence Nationale pour la Recherche (ANR-17-EURE-0017 Frontcog, ANR-17-CE28-0009 GEOMPHON, ANR-10-IDEX-0001-02 PSL*, ANR-19-P3IA-0001 PRAIRIE 3IA Institute, ANR-18-IDEX-0001 U de Paris, ANR-10-LABX-0083 EFL) and grants from CIFAR (Learning in Machines and Brains), Facebook AI Research (Research Grant), Google (Faculty Research Award), Microsoft Research (Azure Credits and Grant), and Amazon Web Service (AWS Research Credits).

\bibliographystyle{acl_natbib}
\bibliography{emnlp2020}

\begin{thebibliography}{22}
\expandafter\ifx\csname natexlab\endcsname\relax\def\natexlab#1{#1}\fi

\bibitem[{Allen and Hospedales(2019)}]{allen}
Carl Allen and Timothy Hospedales. 2019.
\newblock \href {http://proceedings.mlr.press/v97/allen19a.html} {{Analogies
  Explained: Towards Understanding Word Embeddings}}.
\newblock In \emph{Proceedings of the 36th International Conference on Machine
  Learning}, volume~97 of \emph{Proceedings of Machine Learning Research},
  pages 223--231, Long Beach, California, USA. PMLR.

\bibitem[{Bouraoui et~al.(2018)Bouraoui, Jameel, and
  Schockaert}]{bouraoui-etal-2018-relation}
Zied Bouraoui, Shoaib Jameel, and Steven Schockaert. 2018.
\newblock \href {https://www.aclweb.org/anthology/C18-1138} {{Relation
  Induction in Word Embeddings Revisited}}.
\newblock In \emph{Proceedings of the 27th International Conference on
  Computational Linguistics}, pages 1627--1637, Santa Fe, New Mexico, USA.
  Association for Computational Linguistics.

\bibitem[{Chen et~al.(2017)Chen, Peterson, and Griffiths}]{chen-2017}
Dawn Chen, Joshua~C. Peterson, and Thomas~L. Griffiths. 2017.
\newblock \href {http://arxiv.org/abs/1705.04416} {{Evaluating vector-space
  models of analogy}}.
\newblock \emph{CoRR}, abs/1705.04416.

\bibitem[{Devlin et~al.(2019)Devlin, Chang, Lee, and
  Toutanova}]{devlin-etal-2019-bert}
Jacob Devlin, Ming-Wei Chang, Kenton Lee, and Kristina Toutanova. 2019.
\newblock \href {https://doi.org/10.18653/v1/N19-1423} {{{BERT}: Pre-training
  of Deep Bidirectional Transformers for Language Understanding}}.
\newblock In \emph{Proceedings of the 2019 Conference of the North {A}merican
  Chapter of the Association for Computational Linguistics: Human Language
  Technologies, Volume 1 (Long and Short Papers)}, pages 4171--4186,
  Minneapolis, Minnesota. Association for Computational Linguistics.

\bibitem[{Drozd et~al.(2016)Drozd, Gladkova, and
  Matsuoka}]{DrozdGladkovaEtAl_2016_Word_embeddings_analogies_and_machine_learning_beyond_king_man_woman_queen}
Aleksandr Drozd, Anna Gladkova, and Satoshi Matsuoka. 2016.
\newblock \href {https://www.aclweb.org/anthology/C/C16/C16-1332.pdf} {{Word
  Embeddings, Analogies, and Machine Learning: Beyond King - Man + Woman =
  Queen}}.
\newblock In \emph{Proceedings of {{COLING}} 2016, the 26th {{International
  Conference}} on {{Computational Linguistics}}: {{Technical Papers}}}, pages
  3519--3530, {Osaka, Japan, December 11-17}.

\bibitem[{Ethayarajh et~al.(2019)Ethayarajh, Duvenaud, and
  Hirst}]{ethayarajh-etal-2019-towards}
Kawin Ethayarajh, David Duvenaud, and Graeme Hirst. 2019.
\newblock \href {https://doi.org/10.18653/v1/P19-1315} {{Towards Understanding
  Linear Word Analogies}}.
\newblock In \emph{Proceedings of the 57th Annual Meeting of the Association
  for Computational Linguistics}, pages 3253--3262, Florence, Italy.
  Association for Computational Linguistics.

\bibitem[{Finley et~al.(2017)Finley, Farmer, and
  Pakhomov}]{finley-etal-2017-analogies}
Gregory Finley, Stephanie Farmer, and Serguei Pakhomov. 2017.
\newblock \href {https://doi.org/10.18653/v1/S17-1001} {{What Analogies Reveal
  about Word Vectors and their Compositionality}}.
\newblock In \emph{Proceedings of the 6th Joint Conference on Lexical and
  Computational Semantics (*{SEM} 2017)}, pages 1--11, Vancouver, Canada.
  Association for Computational Linguistics.

\bibitem[{Gladkova et~al.(2016)Gladkova, Drozd, and
  Matsuoka}]{GladkovaDrozd2016}
Anna Gladkova, Aleksandr Drozd, and Satoshi Matsuoka. 2016.
\newblock \href {https://doi.org/10.18653/v1/N16-2002} {{Analogy-Based
  Detection of Morphological and Semantic Relations with Word Embeddings: What
  Works and What Doesn't}}.
\newblock In \emph{Proceedings of the NAACL-HLT SRW}, pages 47--54, San Diego,
  California, June 12-17, 2016. ACL.

\bibitem[{Levy and Goldberg(2014)}]{levy-goldberg-2014-linguistic}
Omer Levy and Yoav Goldberg. 2014.
\newblock \href {https://doi.org/10.3115/v1/W14-1618} {{Linguistic Regularities
  in Sparse and Explicit Word Representations}}.
\newblock In \emph{Proceedings of the Eighteenth Conference on Computational
  Natural Language Learning}, pages 171--180, Ann Arbor, Michigan. Association
  for Computational Linguistics.

\bibitem[{Linzen(2016)}]{Linzen_2016_Issues_in_evaluating_semantic_spaces_using_word_analogies}
Tal Linzen. 2016.
\newblock \href {https://doi.org/http://dx.doi.org/10.18653/v1/W16-2503}
  {{Issues in Evaluating Semantic Spaces Using Word Analogies.}}
\newblock In \emph{Proceedings of the {{First Workshop}} on {{Evaluating Vector
  Space Representations}} for {{NLP}}}. {Association for Computational
  Linguistics}.

\bibitem[{Mikolov et~al.(2013{\natexlab{a}})Mikolov, Chen, Corrado, and
  Dean}]{MikolovChenEtAl_2013_Efficient_estimation_of_word_representations_in_vector_space}
Tomas Mikolov, Kai Chen, Greg Corrado, and Jeffrey Dean. 2013{\natexlab{a}}.
\newblock \href {https://arxiv.org/pdf/1301.3781} {{Efficient Estimation of
  Word Representations in Vector Space}}.
\newblock In \emph{Proceedings of {{International Conference}} on {{Learning
  Representations}} ({{ICLR}})}.

\bibitem[{Mikolov et~al.(2013{\natexlab{b}})Mikolov, Yih, and
  Zweig}]{MikolovYihEtAl_2013_Linguistic_Regularities_in_Continuous_Space_Word_Representations}
Tomas Mikolov, Wen-tau Yih, and Geoffrey Zweig. 2013{\natexlab{b}}.
\newblock \href {https://www.aclweb.org/anthology/N13-1090} {Linguistic
  {{Regularities}} in {{Continuous Space Word Representations}}.}
\newblock In \emph{Proceedings of {{NAACL}}-{{HLT}} 2013}, pages 746--751,
  {Atlanta, Georgia, 9\textendash{}14 June 2013}.

\bibitem[{Newman-Griffis et~al.(2017)Newman-Griffis, Lai, and
  Fosler-Lussier}]{newman-griffis-etal-2017-insights}
Denis Newman-Griffis, Albert Lai, and Eric Fosler-Lussier. 2017.
\newblock \href {https://doi.org/10.18653/v1/W17-2303} {{Insights into Analogy
  Completion from the Biomedical Domain}}.
\newblock In \emph{{B}io{NLP} 2017}, pages 19--28, Vancouver, Canada,.
  Association for Computational Linguistics.

\bibitem[{Nissim et~al.(2020)Nissim, van Noord, and van~der
  Goot}]{fairisbetter}
Malvina Nissim, Rik van Noord, and Rob van~der Goot. 2020.
\newblock \href {https://doi.org/10.1162/coli\_a\_00379} {{Fair Is Better than
  Sensational: Man Is to Doctor as Woman Is to Doctor}}.
\newblock \emph{Computational Linguistics}, 46(2):487--497.

\bibitem[{Pennington et~al.(2014)Pennington, Socher, and
  Manning}]{pennington-etal-2014-glove}
Jeffrey Pennington, Richard Socher, and Christopher Manning. 2014.
\newblock \href {https://doi.org/10.3115/v1/D14-1162} {{{G}lo{V}e: Global
  Vectors for Word Representation}}.
\newblock In \emph{Proceedings of the 2014 Conference on Empirical Methods in
  Natural Language Processing ({EMNLP})}, pages 1532--1543, Doha, Qatar.
  Association for Computational Linguistics.

\bibitem[{Radford et~al.(2019)Radford, Wu, Child, Luan, Amodei, and
  Sutskever}]{radford2019language}
Alec Radford, Jeffrey Wu, Rewon Child, David Luan, Dario Amodei, and Ilya
  Sutskever. 2019.
\newblock \href {https://openai.com/blog/better-language-models/} {{Language
  Models are Unsupervised Multitask Learners}}.
\newblock \emph{OpenAI Blog}, 1(8):9.

\bibitem[{Rogers(2019)}]{Rogers_2019_analogies}
Anna Rogers. 2019.
\newblock \href {https://hackingsemantics.xyz/2019/analogies/} {{On word
  analogies and negative results in NLP}}.
\newblock \emph{Hacking Semantics}.

\bibitem[{Rogers et~al.(2017)Rogers, Drozd, and
  Li}]{RogersDrozdEtAl_2017_Too_Many_Problems_of_Analogical_Reasoning_with_Word_Vectors}
Anna Rogers, Aleksandr Drozd, and Bofang Li. 2017.
\newblock \href {http://www.aclweb.org/anthology/S17-1017} {The ({{Too Many}})
  {{Problems}} of {{Analogical Reasoning}} with {{Word Vectors}}}.
\newblock In \emph{Proceedings of the 6th {{Joint Conference}} on {{Lexical}}
  and {{Computational Semantics}} (* {{SEM}} 2017)}, pages 135--148.

\bibitem[{Schluter(2018)}]{schluter-2018-word}
Natalie Schluter. 2018.
\newblock \href {https://doi.org/10.18653/v1/N18-2039} {{The Word Analogy
  Testing Caveat}}.
\newblock In \emph{Proceedings of the 2018 Conference of the North {A}merican
  Chapter of the Association for Computational Linguistics: Human Language
  Technologies, Volume 2 (Short Papers)}, pages 242--246, New Orleans,
  Louisiana. Association for Computational Linguistics.

\bibitem[{Speer et~al.(2017)Speer, Chin, and Havasi}]{speer2017conceptnet}
Robyn Speer, Joshua Chin, and Catherine Havasi. 2017.
\newblock \href {http://aaai.org/ocs/index.php/AAAI/AAAI17/paper/view/14972}
  {{ConceptNet 5.5: An Open Multilingual Graph of General Knowledge}}.
\newblock In \emph{Proceedings of the Thirty-First AAAI Conference on
  Artificial Intelligence}, AAAI’17, page 4444–4451. AAAI Press.

\bibitem[{Tissier et~al.(2017)Tissier, Gravier, and
  Habrard}]{tissier-etal-2017-dict2vec}
Julien Tissier, Christophe Gravier, and Amaury Habrard. 2017.
\newblock \href {https://doi.org/10.18653/v1/D17-1024} {{{D}ict2vec : Learning
  Word Embeddings using Lexical Dictionaries}}.
\newblock In \emph{Proceedings of the 2017 Conference on Empirical Methods in
  Natural Language Processing}, pages 254--263, Copenhagen, Denmark.
  Association for Computational Linguistics.

\bibitem[{Vylomova et~al.(2016)Vylomova, Rimell, Cohn, and
  Baldwin}]{vylomova-etal-2016-take}
Ekaterina Vylomova, Laura Rimell, Trevor Cohn, and Timothy Baldwin. 2016.
\newblock \href {https://doi.org/10.18653/v1/P16-1158} {{Take and Took, Gaggle
  and Goose, Book and Read: Evaluating the Utility of Vector Differences for
  Lexical Relation Learning}}.
\newblock In \emph{Proceedings of the 54th Annual Meeting of the Association
  for Computational Linguistics (Volume 1: Long Papers)}, pages 1671--1682,
  Berlin, Germany. Association for Computational Linguistics.

\end{thebibliography}


\appendix

\section{The offsets' mean direction \label{mean direction}}

For an offsets set, the OCS measures how concentrated the hypercone of the offsets is. The center of this hypercone is the average normalized offset $d$, the unit vector defined by
\setlength{\abovedisplayskip}{3pt}
\setlength{\belowdisplayskip}{2pt}
\begin{equation}
d \propto \sum_i \frac{a_i^* - a_i}{\|a_i^* - a_i\|}
\end{equation}

This vector has different interesting properties. First, we can note that it is the unit vector maximising the sum of the similarities of the offsets with it. With the normalized offsets $o_i\!=\!\frac{a_i^*\!-\!a_i}{\|a_i^*\!-\!a_i\|}$,
\setlength{\abovedisplayskip}{7pt}
\setlength{\belowdisplayskip}{8pt}
\begin{equation}
d = \argmax_{u, \|u\|=1} \sum_i o_i\cdot u
\end{equation}

Furthermore, let's define MSM, the mean similarity of $d$ to the offsets. We can first prove that the MSM is equal to the norm of $d$ before normalization:
\setlength{\abovedisplayskip}{7pt}
\setlength{\belowdisplayskip}{8pt}
 \begin{equation}
 \begin{split}
 \textrm{MSM} &=\frac{1}{N} \sum_j \text{sim}(\frac{\frac{1}{N}\sum_i o_i}{\|\frac{1}{N}\sum_i o_i\|}, o_j) \\ 
 &= \frac{1}{N} \cdot \frac{\sum_i o_i \sum_j o_j}{\|\sum_i o_i\|}=\Big\|\frac{1}{N}\cdot\sum_i o_i\Big\|
 \end{split}
 \end{equation}
 
 Indeed, the average offset will have a norm close to 1 if the offsets are higly similar, but close to 0 if they are anisotropic. We can now go further and link the MSM to the OCS.
\setlength{\abovedisplayskip}{7pt}
\setlength{\belowdisplayskip}{8pt}
 \begin{equation}
 \begin{split}
 \frac{1}{N^2}\Big\|\sum_i o_i\Big\|^2 
 &= \frac{1}{N^2} (\sum_i o_i^2 + \sum_i \sum_{j\neq i} o_i \cdot o_j) \\ 
 &= \frac{1}{N} + \frac{N(N-1)}{N^2}\cdot \textrm{OCS} \\
 \textrm{MSM} &= \sqrt{\frac{1}{N} + \frac{(N-1)}{N}\cdot \textrm{OCS}}
 \end{split}
 \end{equation}
 
\begin{figure*}[ht]
\centering
\includegraphics[width=.8\linewidth]{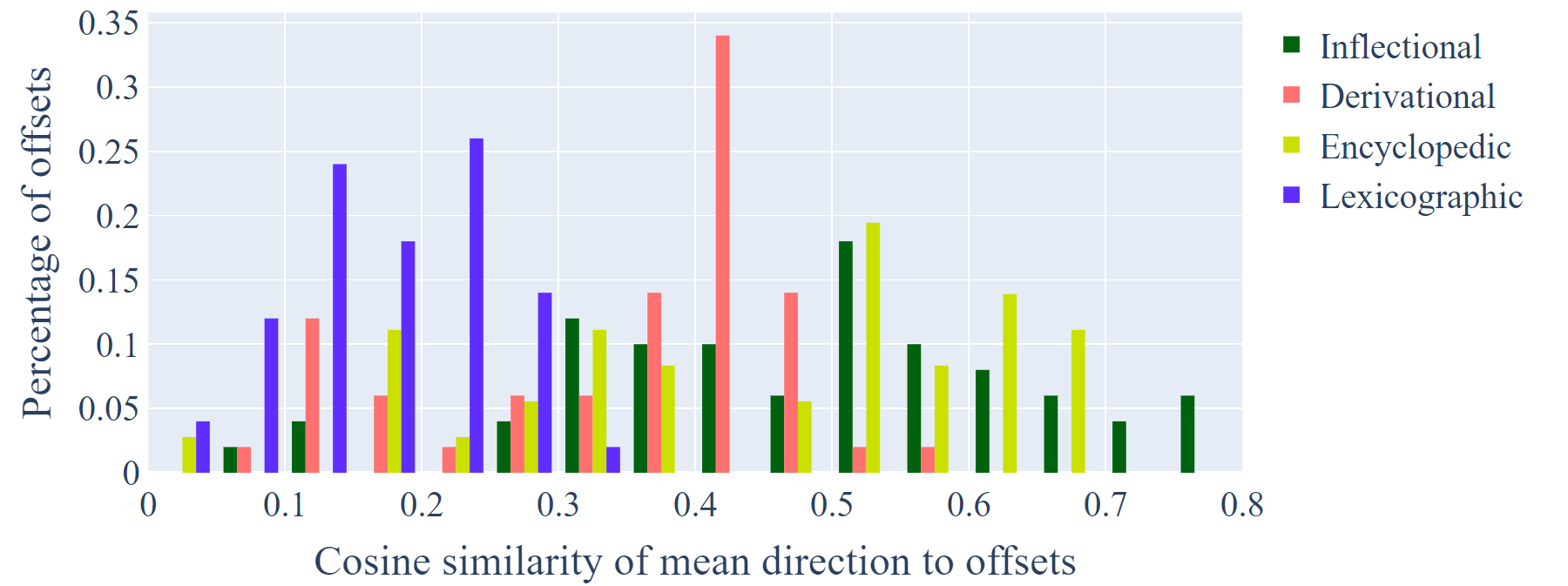}
\caption{Distribution of the cosine similarity of vector offsets to the mean direction for the default word2vec embeddings for each category, then grouped by broad type. The mean similarity improves the similarity even for anisotropic offsets, as discussed. Variance changes seem category dependent.}
\label{fig:distrs to d}
\end{figure*}

 This result shows that the MSM is a strict (parabolic) improvement to the OCS, with a computable baseline decreasing with $N$. For BATS, $N\!=\!50$ (if all words are in the vocabulary), and thus the minimal MSM is $\sqrt{1/50}\!\approx\!0.14$ even for anisotropic offsets (with an $\textrm{OCS}=0$). This score shows that offset concentration can be quickly amplified in unrealistic ways. Still, we can consider $d$ as a close representation of the offsets' relation. We display in Figure \ref{fig:distrs to d} the distribution of the similarities of the offsets to $d$.

\section{Additional figures}

\begin{table*}[ht]
\centering
\scalebox{.9}{\begin{tabular}{l||p{.55cm}p{.55cm}|p{.55cm}p{.55cm}|p{.55cm}p{.55cm}|p{.55cm}p{.55cm}|p{.55cm}p{.55cm}|p{.55cm}p{.55cm}|p{.55cm}p{.55cm}|p{.55cm}p{.55cm}}
 &  \multicolumn{4}{c|}{\textbf{Inflectional}}&\multicolumn{4}{c|}{\textbf{Derivational}}   & \multicolumn{4}{c|}{\textbf{Encyclopedic}}& \multicolumn{4}{c}{\textbf{Lexicographic}} \\
 \textbf{Embedding} 
  &  \multicolumn{2}{c|}{\textbf{Analogy}}  &  \multicolumn{2}{c|}{\textbf{Metrics}} &  \multicolumn{2}{c|}{\textbf{Analogy}}  &  \multicolumn{2}{c|}{\textbf{Metrics}} &  \multicolumn{2}{c|}{\textbf{Analogy}}  &  \multicolumn{2}{c|}{\textbf{Metrics}} &  \multicolumn{2}{c|}{\textbf{Analogy}}  &  \multicolumn{2}{c}{\textbf{Metrics}} \\ \hline
  & N & H & OCS & PCS & N & H & OCS & PCS & N & H &OCS & PCS & N & H &OCS & PCS\\ \hline 
word2vec &  .686
& .099 & .295 & .851  & .158 & .005  & .156 & .679  & .198 & .203 
& .198 & .559 &.063 & .006  & .031 & .539 \\
Glove &  .645 & .224 & .345 & .860 &.167 & .022 & .237 & .710 & .281 
& .122& .255 
& .623 & .080 
& .008 & .004 & .550\\\hline
dict2vec & .503 & .001 & .099 & .700  & .127 & .000 & .079 & .630 & .162 & .050 & .213 & .628 &  .030 & .006 & .024 & .534 \\
Numberbatch & .360 & .226 
& .357 
& .924 
& .060 &.029  & .224 & .805 
& .056  & .066 & .251  & .674 
& .004 & .006 & .034 
& .552 
\\\hline
BERT tokens & .568 & .018  & .217 & .821  &.288 & .229 & .178 & .632 & .041 & .073  & .151 & .569  &  .002 & .015 
& .016 & .517 \\
GPT-2 tokens & .119 & .019 & .097 & .651  &.359 
& .248 
& .270 
& .732 & .014 & .050 & .071 & .513  &  .003 & .011 & .011 & .508\\
\end{tabular}}
\caption{Comparison of analogy test scores and our OCS and PCS metrics for different word embeddings, grouped by broad type. N designates the Normal analogy test scores, and H the Honest analogy test scores where the input words are allowed to be predicted. OCS and PCS designate the Offsets Concentration Scores and Pairwise Consistency Scores. Bold means highest score for this type.}
\label{tab analogy embeddings results annex}
\end{table*}

\begin{figure*}[ht]
  \centering
\includegraphics[width=.9\linewidth]{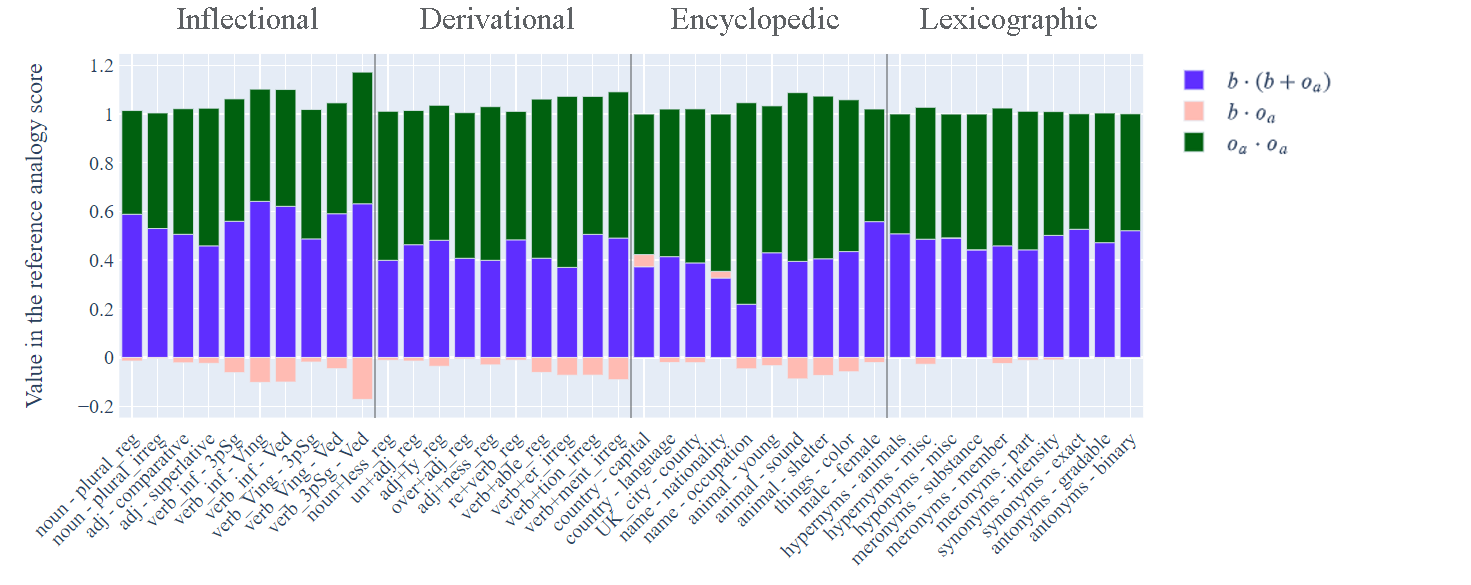}
  \caption{Decomposition of the analogy score if $o_a\!=\!o_b$ (the score being therefore equal to 1). The similar values of $b\cdot (b+o_a)$ and $o_a \cdot o_a$ indicates that the length of $o_a$ and $b$ is similar, and thus that the offset length does not explain the low value of $o_a\!\cdot\!o_b$ in the real decomposition.} 
  \label{fig:bar decompo ref}
\end{figure*}

\end{document}